\documentclass{article}

\usepackage{microtype}
\usepackage{graphicx}
\usepackage{subcaption}
\usepackage{booktabs} 

\usepackage{hyperref}

\usepackage[accepted]{icml2026}

\usepackage{amsmath}
\usepackage{amssymb}
\usepackage{mathtools}
\usepackage{amsthm}
\usepackage{threeparttable}
\usepackage{pifont}
\usepackage{xcolor}
\usepackage{makecell}
\usepackage{listings}
\usepackage{lipsum}
\usepackage{multirow}
\usepackage{booktabs}
\usepackage{graphicx}

\usepackage[capitalize,noabbrev]{cleveref}

\theoremstyle{plain}

\theoremstyle{definition}

\theoremstyle{remark}

\usepackage[textsize=tiny]{todonotes}

\icmltitlerunning{Fine-tuning DeepSeek-OCR-2 for Molecular Structure Recognition}

\begin{document}

\twocolumn[
  \icmltitle{Fine-tuning DeepSeek-OCR-2 for Molecular Structure Recognition}



  \icmlsetsymbol{equal}{*}

  \begin{icmlauthorlist}
    \icmlauthor{Haocheng Tang}{pitt,ccgs,neu}
    \icmlauthor{Xingyu Dang}{pton}
    \icmlauthor{Junmei Wang}{pitt,ccgs}
  \end{icmlauthorlist}

  \icmlaffiliation{pton}{Department of Computer Science, Princeton University, Princeton, NJ, USA}
  \icmlaffiliation{pitt}{School of Pharmacy, University of Pittsburgh, Pittsburgh, PA, USA}
  \icmlaffiliation{neu}{Khoury College of Computer Science, Northeastern University, Boston, MA, USA}
  \icmlaffiliation{ccgs}{Computational Chemical Genomics Screening Center, University of Pittsburgh, Pittsburgh, PA, USA}

  \icmlcorrespondingauthor{Haocheng Tang}{tang.haoc@northeastern.edu}
  \icmlcorrespondingauthor{Junmei Wang}{juw79@pitt.edu}

  \icmlkeywords{Large Language Models, Machine Learning, Chemical Reactions, Chemistry}

  \vskip 0.3in
]



\printAffiliationsAndNotice{}  

\begin{abstract}

Optical Chemical Structure Recognition (OCSR) is critical for converting 2D molecular diagrams from printed literature into machine-readable formats. While Vision-Language Models have shown promise in end-to-end OCR tasks, their direct application to OCSR remains challenging, and direct full-parameter supervised fine-tuning often fails. In this work, we adapt DeepSeek-OCR-2 \cite{wei2025deepseek,wei2026deepseek2} for molecular optical recognition by formulating the task as image-conditioned SMILES generation. To overcome training instabilities, we propose a two-stage progressive supervised fine-tuning strategy: starting with parameter-efficient LoRA and transitioning to selective full-parameter fine-tuning with split learning rates. We train our model on a large-scale corpus combining synthetic renderings from PubChem and realistic patent images from USPTO-MOL to improve coverage and robustness. Our fine-tuned model, MolSeek-OCR, demonstrates competitive capabilities, achieving exact matching accuracies comparable to the best-performing image-to-sequence model. However, it remains inferior to state-of-the-art image-to-graph modelS. Furthermore, we explore reinforcement-style post-training and data-curation-based refinement, finding that they fail to improve the strict sequence-level fidelity required for exact SMILES matching.
 
\end{abstract}

\section{Introduction}

The automated extraction of chemical knowledge relies heavily on the accurate digitization of complex scientific documents. A critical component of this digitization process is Optical Chemical Structure Recognition (OCSR), which aims to convert 2D molecular diagrams from printed literature into machine-readable formats, such as SMILES or molecular graphs. Historically, deep learning models for OCSR have evolved from Transformer-based encoder-decoder architectures that translate image pixels directly into SMILES strings, such as DECIMER \cite{rajan2023decimer}, to more advanced approaches like MolScribe \cite{qian2023molscribe} that explicitly predict atoms, bonds, and their geometric layouts to construct robust 2D graphs.

Concurrently, the rapid development of Large Language Models (LLMs) and Vision-Language Models (VLMs) has transformed how AI interacts with chemical data. To overcome the limitations of text-only representations, multimodal LLMs have incorporated structured molecular inputs to enable joint reasoning over language and molecular structures. In the broader domain of document understanding, fine-tuning VLMs for OCR has increasingly shifted towards end-to-end, OCR-free architectures, eliminating the reliance on pipeline-based OCR engines. These fine-tuned document VLMs provide the essential text and layout grounding required for multimodal chemistry models to autonomously ingest and reason over unstructured scientific literature.

Building on these parallel advancements in OCSR and document VLMs, this work adapts DeepSeek-OCR-2 to molecular optical recognition by formulating the task as image-conditioned SMILES generation. Because direct full-parameter supervised fine-tuning (SFT) failed, we introduce a two-stage progressive supervised fine-tuning strategy. Our approach begins with parameter-efficient fine-tuning using LoRA to adapt the text generation pathway and cross-modal alignment interface. Subsequently, we apply progressive full-parameter fine-tuning, freezing the lowest-level visual tokenizer while optimizing higher-level modules using split learning rates for the visual and language branches.

To ensure robustness across diverse drawing styles and document artifacts, we utilize a mixed training dataset combining large-scale synthetic data from PubChem with realistic patent images from USPTO-MOL \cite{Rajan2020uspto, Staker2019staker,Kim2016pubchem,Yoo2022i2g}. Finally, we benchmark our resulting models against traditional baselines to evaluate their exact matching accuracy across synthetic, realistic, and perturbed datasets. We also explore the limitations of reinforcement-style post-training methods, such as Group Sequence Policy Optimization (GSPO) and Representation Finetuning (ReFT), highlighting the unique challenges of maintaining strict sequence-level fidelity in VLM-based molecular recognition.

\section{Related Work}

\subsection{LLMs for Chemistry}

LLMs are increasingly utilized in chemistry for molecular understanding and scientific knowledge grounding. Initial applications relied heavily on prompt engineering, proving that general-purpose models could tackle chemical problems without parameter updates \cite{Hatakeyama-Sato2023pegpt4, Liu2024pe,tang2025adseqgan}. This success inspired chemistry-specific fine-tuned models—such as ChemLLM \cite{zhang2024chemllm}, ChemDFM \cite{zhao2025developing}, and BatGPT-Chem \cite{Yang2025batgptchem}—which adapt robust pretrained backbones to specialized chemical corpora. However, recognizing the limitations of text-only data for complex structural parsing, recent multimodal models have incorporated structured inputs like SMILES, 2D graphs, and 3D geometries. Frameworks like nach0 \cite{Livne2024nach0}, InstructMol \cite{cao2023instructmol}, Chem3DLLM \cite{jiang2025chem3dllm}, and ChemVLM \cite{li2025chemvlm} exemplify this shift, laying the essential foundation for joint visual-linguistic reasoning over molecular structures.

\subsection{Molecular Structure Recognition}

OCSR is critical for converting 2D molecular diagrams from printed literature into machine-readable formats like SMILES or molecular graphs. While early deep learning models such as DECIMER \cite{rajan2023decimer} utilized Transformer-based encoder-decoder architectures to translate image pixels directly into SMILES strings, subsequent approaches like MolScribe \cite{qian2023molscribe} advanced the field by explicitly predicting atoms, bonds, and their geometric layouts to construct robust 2D graphs. More recently, VLMs have been adapted for OCSR. To overcome the limitations of treating molecular recognition merely as a holistic image-captioning task, cutting-edge frameworks like GTR-VL \cite{gtrvl2025} introduce a "Visual Chain of Thought" mechanism. This approach mimics human reasoning by incrementally parsing molecular graphs through sequential atom-bond predictions, significantly improving the recognition of complex and hand-drawn structures.

\subsection{Fine-tuning VLMs for OCR Tasks}

The automated extraction of chemical knowledge relies heavily on the accurate digitization of complex scientific documents. Consequently, fine-tuning VLMs for Optical Character Recognition (OCR) has increasingly shifted towards end-to-end, OCR-free architectures. Models like Donut \cite{kim2022donut} eliminate the reliance on pipeline-based OCR engines by directly mapping document images to structured text using a vision encoder and text decoder. For scientific literature specifically, Nougat \cite{blecher2023nougat} adapts this paradigm to convert dense academic PDFs—including tables, inline math, and chemical equations—directly into Markdown markup. These fine-tuned document VLMs provide the essential text and layout grounding required for multimodal chemistry models to autonomously ingest and reason over unstructured scientific literature.

\section{Methodology}

\subsection{Dataset}

\begin{table}
\centering
\caption{Summary of the Benchmarks for Evaluation}
\resizebox{\columnwidth}{!}{
\begin{tabular}{llcc}
\toprule
Dataset & Source & No. of Images & \% of chiral \\
\midrule
Indigo & synthetic & 5,719 & 20.2\% \\
ChemDraw & synthetic & 5,719 & 20.2\% \\
CLEF & patent (US) & 992 & 32.7\% \\
JPO & patent (Japanese) & 450 & 0\% \\
UOB & catalog & 5,740 & 0\% \\
USPTO & patent (US) & 5,719 & 20.2\% \\
Staker & patent (US) & 50,000 & 17.3\% \\
ACS & journal article & 331 & 19.3\% \\
\bottomrule
\end{tabular}
}
\end{table}

\begin{figure*}[!h]
    \centering
    \includegraphics[width=1\linewidth]{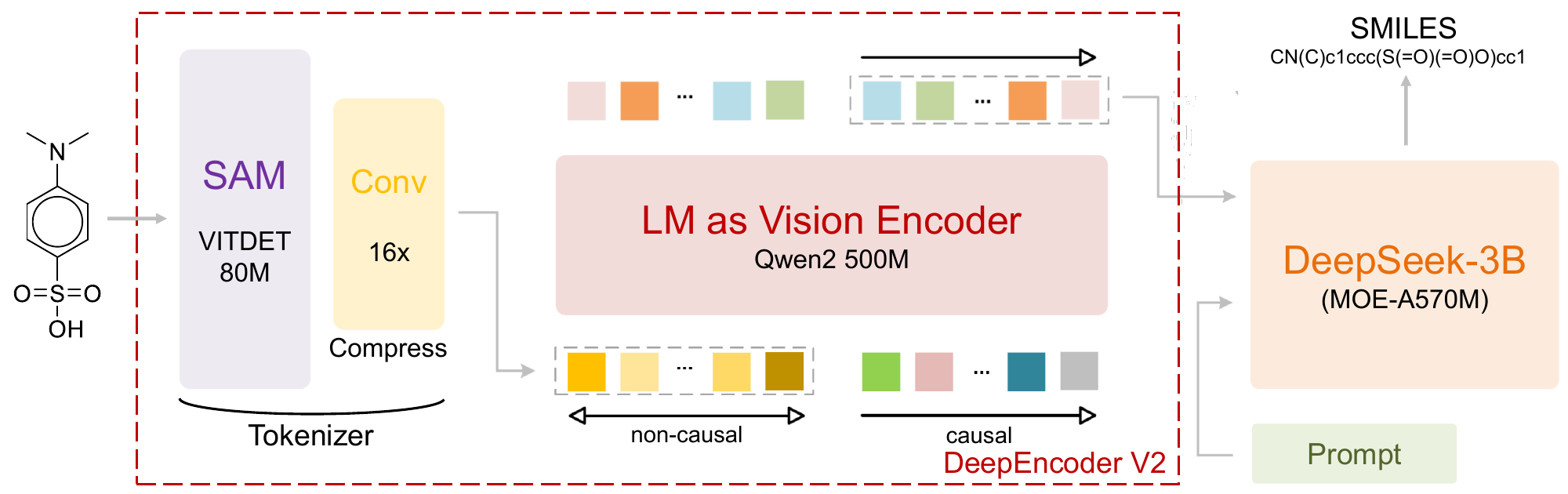}
    \caption{\textbf{The overall architecture of DeepSeek-OCR-2.}}
    \label{fig:1}
\end{figure*}

\paragraph{Training}
Following the common practice in molecular structure recognition, we combine large-scale synthetic data with realistic patent images to improve both coverage and robustness following Molscribe. Our training corpus is built from two sources. The first source is PubChem\cite{Kim2016pubchem}, which provides a large collection of molecular structures paired with SMILES strings. We use these structures to render molecular depictions on the fly, allowing the model to observe diverse drawing styles during training. In particular, we generate two complementary synthetic domains: a MolScribe-like style with stronger appearance variation and a cleaner ChemDraw-like style with fewer perturbations. This synthetic branch improves the model's tolerance to style diversity, bond rendering differences, and local annotation variation. The second source is USPTO-MOL \cite{qian2023molscribe}, which contains realistic molecular images extracted from patents together with paired structure annotations. This branch exposes the model to real document artifacts, including imperfect scan quality, nonuniform line thickness, and patent-specific drawing conventions.

The two training stages use the same data sources but different sampling budgets. In the LoRA stage, we sample $64$k examples from each of the three subsets, namely PubChem rendered in the MolScribe-like style, PubChem rendered in the ChemDraw-like style, and realistic USPTO-MOL images, resulting in $192$k training instances in total. In the subsequent progressive full fine-tuning stage, we enlarge the corpus to $800$k examples, consisting of $300$k MolScribe-like rendered molecules from PubChem, $300$k ChemDraw-like rendered molecules from PubChem, and $200$k realistic molecular images from USPTO-MOL. Across all subsets, the supervision target is the corresponding SMILES sequence.

\paragraph{Evaluation}

We evaluate the model on both clean and perturbed benchmarks that cover synthetic renderings, patent images, catalog images, and journal figures \cite{Rajan2020uspto, Staker2019staker}. The clean benchmark suite contains two synthetic sets, namely ChemDraw and Indigo, and five realistic sets, namely USPTO, CLEF, Staker, UOB, and ACS. The two synthetic sets measure in-domain recognition under standardized rendering engines, whereas the realistic sets evaluate generalization to authentic molecular depictions collected from patents, catalogs, and journal articles. In addition, we report robustness on four perturbation sets derived from CLEF, Staker, UOB, and USPTO, where the input images are further corrupted by appearance variations and quality degradations. This evaluation protocol reflects the central challenge of OCSR: strong performance is required not only on clean synthetic drawings but also on heterogeneous real-world images and visually perturbed inputs.

\subsection{Training Pipeline}

We adapt DeepSeek-OCR-2 to molecular optical recognition by formulating the task as image-conditioned SMILES generation. Given a molecular depiction and a fixed instruction prompt, the model autoregressively predicts the target SMILES sequence. Our training recipe follows a two-stage progressive supervised fine-tuning strategy. Direct full-parameter SFT failed. In both stages, we preserve the original multi-scale visual pipeline of DeepSeek-OCR-2: each sample is processed with one global view and optional local crops so that both overall molecular topology and local bond-level details can be retained. Supervision is applied only to the generated answer tokens, namely the target SMILES string, rather than to the prompt or image placeholder tokens.

\paragraph{Architecture of DeepSeek-OCR-2}
As illustrated in Figure~\ref{fig:1}, DeepSeek-OCR-2 consists of three main components: a visual tokenizer, an LM-as-vision-encoder module, and an autoregressive decoder, with a compression/projection interface connecting the visual and language streams. The visual tokenizer converts the molecular image into a compact sequence of visual tokens, while the LM-style visual encoder reorganizes and refines these tokens before passing them to the decoder. This design is well suited to molecular OCR because it combines fine-grained visual perception with sequential chemical string generation. In particular, the encoder is able to preserve structural cues such as rings, branches, and bond patterns, whereas the decoder maps the reordered visual representation into a valid SMILES sequence.

\begin{table*}[t]
\centering
\caption{Exact matching accuracy (\%) on different datasets.}
\resizebox{\textwidth}{!}{
\begin{tabular}{llccccccccccc}
\toprule
 &  & \multicolumn{3}{c}{Synthetic} & \multicolumn{4}{c}{Realistic} & \multicolumn{4}{c}{Perturbed} \\
\cmidrule(lr){3-5} \cmidrule(lr){6-9} \cmidrule(lr){10-13}
Models & Method 
& Indigo & ChemDraw & CLEF 
& UOB & USPTO & Staker & ACS 
& CLEF$_p$ & UOB$_p$ & USPTO$_p$ & Staker$_p$ \\
\midrule

\multirow{2}{*}{Rule-based}
& MolVec 
& 95.4 & 87.9 & 82.8 
& 80.6 & 88.4 & 0.8 & 47.4 
& 43.7 & 74.5 & 29.7 & 5.0 \\
& OSRA 
& 95.0 & 87.3 & 84.6 
& 78.5 & 87.4 & 0.0 & 55.3 
& 11.5 & 68.3 & 4.0 & 4.6 \\

\midrule

\multirow{3}{*}{Image to Sequence}
& Img2Mol 
& 58.9 & 46.4 & 18.3
& 68.7 & 26.3 & 17.0 & 23.0 
& 21.1 & 74.9 & 29.7 & 51.7 \\
& DECIMER 
& 69.6 & 86.1 & 62.7 
& \textbf{88.2} & 41.1 & 40.8 & 46.5 
& 70.6 & \textbf{87.3} & 46.4 & 47.9 \\
& SwinOCSR 
& 74.0 & 79.6 & 30.0 
& 44.9 & 27.9 & -- & 27.5 
& 32.2 & -- & -- & -- \\

\midrule

\multirow{4}{*}{Image To Graph}
& MSE-DUDL 
& -- & -- & --
& -- & -- & 77.0 & -- 
& -- & -- & -- & -- \\
& ChemGrapher 
& -- & -- & -- 
& 70.6 & -- & -- & -- 
& -- & -- & -- & -- \\
& Image2Graph 
& -- & -- & 51.7 
& 82.9 & 55.1 & -- & -- 
& -- & -- & -- & -- \\
& MolScribe 
& \textbf{97.5} & \textbf{93.8} & \textbf{88.9} 
& 87.9 & \textbf{92.6} & \textbf{86.9} & \textbf{71.9} 
& \textbf{90.4} & 86.7 & \textbf{92.5} & 65.0 \\

\midrule
\multirow{3}{*}{VLM-based}
& DeepSeek-OCR-2 
& 0.0 & 0.0 & 0.0
& 0.2 & 0.0 & 0.0 & 0.0 
& 0.0 & 0.1 & 0.0 & 0.0 \\
& MolSeek-OCR-LoRA
& 68.4 & 65.1 & 54.8 
& 64.3 & 50.6 & 46.3 & 24.8 
& 62.0 & 44.0 & 56.2 & 30.0 \\
& MolSeek-OCR 
& 74.3 & 72.2 & 63.3
& 72.6 & 61.0 & 50.5 & 29.9
& 70.7 & 49.7 & 65.6 & 31.2 \\

\bottomrule
\end{tabular}
}
\footnotesize{
$^a$ Scores are exact matching accuracy. \\
$^b$ Results from the original papers; -- means not available. \\
Img2Mol does not predict chirality. Additional evaluation ignoring chirality is in the Supporting Information. \\
$^d$ SwinOCSR results on large datasets are omitted due to long runtime.
}
\end{table*}

\paragraph{LoRA SFT}

\begin{figure*}[!h]
    \centering
    \includegraphics[width=0.8\linewidth]{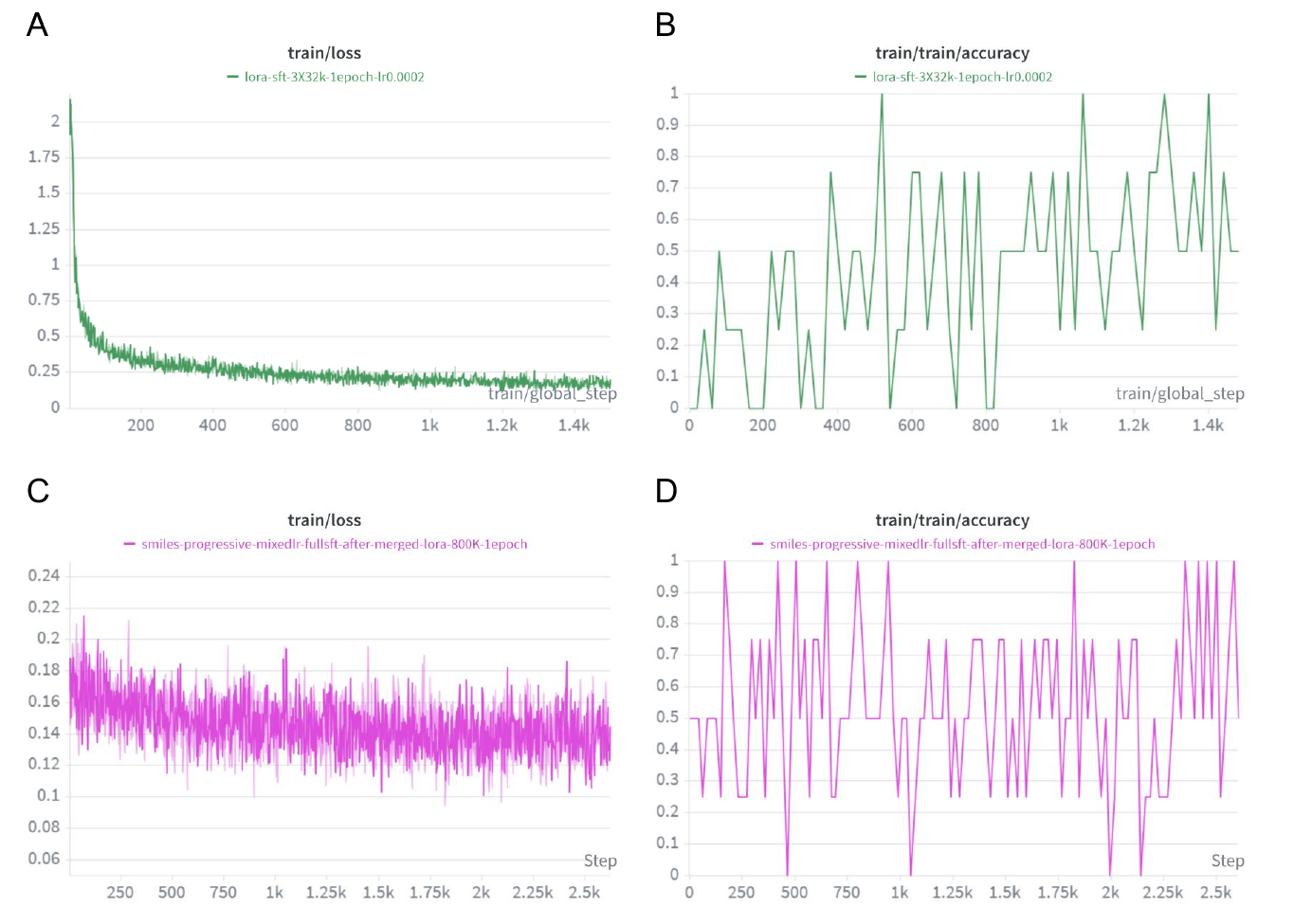}
    \caption{\textbf{The 2 stages fine-tuning.} A. Train loss in stage 1. B. Train accuracy in stage 2. C. Train loss in stage 2. C. Train accuracy in stage 2. }
    \label{fig:1}
\end{figure*}

In the first stage, we perform parameter-efficient supervised fine-tuning with LoRA starting from the released DeepSeek-OCR-2 checkpoint. Rather than adapting only the final decoder, we place LoRA modules on the main attention and feed-forward projections, as well as on the visual-language projection layers highlighted by Figure~\ref{fig:1}. Consequently, this stage adapts not only the text generation pathway but also the alignment interface between the visual encoder and the decoder, which is important for translating molecular drawings into chemical strings.

The LoRA stage uses a three-source mixed training set with $32$k samples drawn from each source, resulting in $96$k training instances in total. Two subsets are constructed from PubChem molecules using on-the-fly rendering. One subset follows a MolScribe-like style with stronger appearance augmentation, while the other follows a cleaner ChemDraw-like style with less augmentation. A third subset is built from realistic molecule images from USPTO-MOL. For synthetically rendered samples, the supervision target is the SMILES string corresponding to the rendered depiction; for realistic samples, the target is the paired SMILES annotation provided with the image. All samples are trained with the same instruction asking the model to return the SMILES of the input molecule.

\paragraph{Progressive full SFT}
In the second stage, we continue training from the LoRA-adapted checkpoint and switch to progressive full-parameter fine-tuning. The training set is expanded to $800$k samples, including $300$k MolScribe-like rendered molecules from PubChem, $300$k ChemDraw-like rendered molecules from the same source, and $200$k realistic molecular images from USPTO-MOL. The prompt format and the response-only objective remain unchanged.

This stage does not update the entire network uniformly. Instead, we freeze the lowest-level visual tokenizer in Figure~\ref{fig:1} together with the input token embedding layer, and continue optimizing the higher-level LM-as-vision-encoder module, the compression/projection interface, and the autoregressive decoder. We further use split learning rates for different branches: a smaller rate for the visual branch and a larger rate for the language generation branch. This design reflects the intuition that low-level visual perception is already reasonably stable after the first stage, whereas higher-level cross-modal alignment and sequence generation require more task-specific adaptation for molecular OCR.

\section{Experiment}

\subsection{Performance of SFT models}

As detailed in Table 2, the fine-tuned VLM models demonstrate competitive chemical structure recognition capabilities compared to traditional baselines. Notably, the full-parameter fine-tuned MolSeek-OCR achieves exact matching accuracies that are broadly comparable to DECIMER, the best-performing Image-to-Sequence model, across various synthetic, realistic, and perturbed datasets. However, despite these improvements over the zero-shot capabilities of the base model, the VLM-based approaches are still inferior to state-of-the-art Image-to-Graph models. MolScribe consistently outperforms the SFT VLMs by a significant margin, suggesting that explicitly predicting geometric node-and-edge layouts currently retains an advantage over autoregressive text generation for complex structural parsing. MolScribe and GTR-VL \cite{gtrvl2025} are widely regarded as state-of-the-art methods in this area.
 
\subsection{Failure of GSPO and ReFT}

Beyond progressive SFT, we also explored reinforcement-style post-training with GSPO and data-curation-based refinement with ReFT. In both cases, the starting point was the progressively fine-tuned model. The overall pipeline was to first sample multiple candidate SMILES sequences for each molecular image, then score these candidates with a chemistry-aware reward, and finally use the selected signals to further update the model. The reward was designed to jointly encourage chemical validity, sequence-level correctness, graph-level consistency after removing chirality-sensitive distinctions, and better handling of stereochemical cases. We also tested a broad range of training settings and reward trade-offs during this stage, aiming to balance exploration with stable optimization.

However, these attempts did not lead to a useful improvement over progressive SFT. Across our experiments, we consistently observed that the exact-match accuracy of SMILES tended to decrease during the intermediate optimization process, even when the graph-level accuracy showed a small improvement. This pattern suggests that the additional optimization encouraged the model to move toward graph-equivalent or nearly correct structures, but often at the expense of the precise serialized SMILES form required by the benchmark. In other words, the reward signal was able to recover part of the underlying molecular graph, yet it was insufficient to preserve the strict sequence-level fidelity needed for exact matching. Since our primary evaluation protocol emphasizes exact molecular reconstruction, and the net effect on final benchmark performance was unfavorable, we did not adopt GSPO or ReFT in the final training recipe.

\section{Conclusion}

Through a two-stage progressive fine-tuning strategy, we successfully established a stable transfer path from a document OCR foundation model to molecule-to-SMILES recognition. The resulting model, MolSeek-OCR, demonstrates competitive capabilities, achieving exact matching accuracies that are broadly comparable to DECIMER, the best-performing Image-to-Sequence baseline. Despite these advancements, our VLM-based approach remains inferior to state-of-the-art Image-to-Graph models like MolScribe, indicating that explicitly predicting geometric node-and-edge layouts is currently superior to autoregressive text generation for structural parsing. Additionally, our explorations into reinforcement-style post-training and data-curation-based refinement did not yield useful improvements. We observed that these intermediate optimization processes encouraged graph-equivalent structures but failed to preserve the strict sequence-level fidelity required for exact SMILES matching.

\section*{Software and Data}

All the training datasets, code and parameters are available online at GitHub: \texttt{https://github.com/HaCTang/MolSeek-OCR}.

\section*{Acknowledgements}

This work was supported by funds from the National Institutes of Health (R01GM147673, R01GM149705) and the National Science Foundation (1955260). The authors would like to thank the computing resources provided by the Center for Research Computing (facility RRID: SCR\_022735) at the University of Pittsburgh (NSF award number OAC-2117681), and the Pittsburgh Supercomputer Center (grant number BIO210185)

\nocite{langley00}

\bibliography{example_paper}
\bibliographystyle{icml2026}

\newpage
\appendix
\onecolumn
\section{Training Details}

\subsection{LoRA SFT}
We set the LoRA rank to 64, the scaling factor to 64, and the dropout rate to 0.1. Optimization uses AdamW with a learning rate of $2\times 10^{-4}$, weight decay of $10^{-3}$, 100 warmup steps, batch size 4 per device, and gradient accumulation of 8. Training runs for 3000 update steps.

\subsection{Full SFT}

Optimization in this stage uses AdamW with a base learning rate of $1\times 10^{-5}$, reduced to $5\times 10^{-6}$ on the visual branch and increased to $2\times 10^{-5}$ on the language branch, together with weight decay of $0.01$, 250 warmup steps, batch size 4 per device, and gradient accumulation of 8. We train for 2500 update steps. Overall, this two-stage strategy first identifies an efficient adaptation subspace through LoRA and then performs selective full fine-tuning on semantically higher-level modules, yielding a stable transfer path from a document OCR foundation model to molecule-to-SMILES recognition.


\end{document}